\documentclass[10pt, a4paper]{article}
\usepackage{lrec}
\usepackage{graphicx}
\usepackage{tabularx}
\usepackage{booktabs}
\usepackage{soul}
\usepackage{amsmath}
\usepackage{amsfonts}
\usepackage{tcolorbox}
\usepackage{xcolor}
\newcommand*{\mybox}[1]{\framebox{#1}}
\usepackage{epstopdf}
\usepackage[latin1]{inputenc}
\usepackage{enumitem}

\usepackage{hyperref}
\usepackage{xstring}
\usepackage{amsthm}

\theoremstyle{definition}
\newtheorem{definition}{Definition}

\title{Natural Language Premise Selection: Finding Supporting Statements for Mathematical Text}

\name{Deborah Ferreira and Andre Freitas}

\address{Department of Computer Science \\
         University of Manchester \\
         \{deborah.ferreira, andre.freitas\}@manchester.ac.uk}

\abstract{Mathematical text is written using a combination of words and mathematical expressions. This combination, along with a specific way of structuring sentences makes it challenging for state-of-art NLP tools to understand and reason on top of mathematical discourse. In this work, we propose a new NLP task, the natural premise selection, which is used to retrieve supporting definitions and supporting propositions that are useful for generating an informal mathematical proof for a particular statement. We also make available a dataset, NL-PS, which can be used to evaluate different approaches for the natural premise selection task. Using different baselines, we demonstrate the underlying interpretation challenges associated with the task.
\\ \newline \Keywords{mathematical text, mathematical language processing, mathematical text analysis} }
\begin{document}

\maketitleabstract

\section{Introduction}

Comprehending mathematical text requires evaluating the semantics of its mathematical structures (such as expressions) and connecting its internal components with the respective definitions or premises~\cite{Greiner-Petter2019WhyYet}.

State-of-the-art models for natural language processing, such as BERT~\cite{devlin2018bert}, have high scores for several tasks, such as entity recognition, textual entailment and machine translation, but they do not encode the intricate mathematical background knowledge needed to reason over mathematical discourse. 

The language of mathematics is composed of a combination of words and symbols, where symbols follow a different set of rules and have a specific alphabet. Nonetheless, word and symbols are interdependent in the context of mathematical discourse. This phenomenon is exclusive to mathematical language, not found in any other natural, or artificial, language~\cite{ganesalingam2013language}, providing a unique and challenging application for semantic evaluation and natural language processing.

Understanding mathematical discourse has been explored before as a Mathematical Knowledge Extraction task~\cite{aizawa2014ntcir}; however, several aspects of the mathematical discourse related to deeper and more granular reasoning over mathematical discourse has not yet been investigated. There is a lack of datasets in the literature needed for exploring and studying mathematical discourse and its associated interpretation and reasoning.

We propose the task of natural premise selection, inspired by the field of automated theorem processing. Premise selection appeared initially as a task of selecting a (useful) part of an extensive formal library in order to limit the search space for an \textit{Automated Theorem Proving} (ATP) system, increasing the chance of finding a proof for a given conjecture~\cite{blanchette2016hammering}. Premises considered relevant are the ones that ATPs use for the automatic deduction process of finding a proof for a conjecture. The premise selection task is defined as: Given a collection of premises $P$, an ATP system $A$ with given resource limits, and a new conjecture $c$, predict those premises from $P$ that will most likely lead to an automatically constructed proof of $c$ by $A$~\cite{irving2016deepmath}. 

Natural premise selection is based not on formally structured mathematics, but on human-generated mathematical text. It takes as input mathematical text, written in natural language and outputs relevant mathematical statements that could support a human in finding a proof for that mathematical text. The premises are composed by a set of supporting definitions and supporting propositions, that act as explanations for the proof process.

For example, the famous \textit{Fermat's Little Theorem}~\cite{warner1990modern} has different possible proofs, one of them using the \textit{Euclid's Lemma}. In this example, Euclid's Lemma would be considered useful for a human trying to prove Fermat's Little Theorem; therefore, it is a premise for the conjecture that Fermat's Little Theorem presents.

In order to evaluate this task, we propose a new dataset: NL-PS (Natural Language - Premise Selection), using as a basis the human-curated website ProofWiki\footnote{https://proofwiki.org/wiki/Main\_Page}. This dataset opens possibilities of applications not only for the premise selection task but also for evaluating semantic representations for mathematical discourse (including embeddings), textual entailment for mathematics and natural language inference in the context of mathematical texts. 

The contributions of this paper can be summarised as follows:
\begin{itemize}
    \item Proposal of a new NLP task: natural language premise selection.
    \item A novel dataset, NL-PS, to support the evaluation of premise selection methods using natural language corpora. 
    \item Comparison of different baselines for the natural premise selection task.
\end{itemize}

\section{Related Work}

NLP has been applied before in the context of general Mathematics. \newcite{chaganty2016much} proposes a new task for semantic analysis, the task of perspective generation, i.e., generating description to numerical values using other values as reference. \newcite{huang2016well} analyze different approaches to solve mathematical word problems and concludes that it is still an unsolved challenge. 

\newcite{ganesalingam2017fully} propose a program that solves elementary mathematical problems, mainly in metric space theory, and presents solutions similar to the ones proposed by humans. The authors recognize that their system is operating at a disadvantage because human language involves several constraints that rule out many sound and effective tactics for generating proofs. 

\newcite{wang2018first} propose an approach to automatically formalize informal mathematics using statistical parsing methods and large-theory automated reasoning. The idea is to convert from an informal statement to a formal one, using Mizar as the output language. After the statement has been correctly translated, it can be checked using an automatic tool.

Naproche (Natural language Proof Checking)~\cite{cramer2009naproche} is a project focused on the development of a controlled natural language (CNL) for mathematical texts and adapting proof checking software to work with this language in order to check syntactical and mathematical correctness.

\newcite{zinn2003computational} proposes proof representation structures to represent mathematical discourse using discourse representation theory and also proposes a prototype that could be used to automate the process of generating proofs.

Approaches for creating embeddings of mathematical text have applied variations of the Skip-gram model~\cite{mikolov2013distributed}, extending it with a specific tokenization strategy for equations and mathematical terms. Most tokenization strategies will use the tree structure of an equation to define the target tokens and can range from considering the full equation~\cite{Krstovski2018EquationEmbeddings} as a single token or decomposing its component expressions or at the individual symbol-level~\cite{DBLP:journals/corr/GaoJYYYT17}.
\newcite{Greiner-Petter2019WhyYet} developed a skip-gram-based model using as a reference corpus a set of arXiv papers in HTML format using a term-level tokenization granularity. The authors found that the induced vector space did not produce meaningful semantic clusters.

Premise selection is an approach generally used for selecting useful premises to prove conjectures in Automated Theorem Proving (ATP) systems~\cite{alama2014premise}. \newcite{irving2016deepmath} propose a neural architecture for premise selection using formal statements written in Mizar. Other authors have used machine learning approaches such as Kernel-based Learning~\cite{Alama2014}, k-NN algorithm~\cite{Gauthier:2015:PSE:2676724.2693173} and Random Forests~\cite{10.1007/978-3-319-24246-0_20}. However, the neural approaches previously presented~\cite{irving2016deepmath} have obtained higher scores at the premise selection task.

\section{Linguistic Considerations}

In this section, we describe some of the linguistics features present in a mathematical corpus. Our aim is to examine its discourse in combination with natural language. The following definitions are not of mathematical objects since those already have established mathematical definitions; in this work, we are interested in how the different mathematical objects are presented inside the mathematical text.

\begin{definition}A \textbf{mathematical expression} $\mathcal{M}$, in a mathematical text, is defined by a set $\Sigma = \{s_1, s_2, s_3, ..., s_n\}$ where $s_i \in S$, and $S$ is the set of symbols present in a certain mathematical domain of discourse, such as variables, constants and functions. A variable, for example, is considered an expression.
\end{definition}

\begin{definition}An \textbf{equation} $\mathcal{E}$ is defined as a combination of $m_i, m_j \in \mathcal{M}$ and an (in)equality predicate $p \in \{>, <, \leq, \geq, \neq, = \}$. 
\end{definition}

\begin{definition}A \textbf{mathematical statement} $\mu$ can be:
\begin{itemize}[noitemsep]
    \item A sequence of words (from the mathematical domain) or;
    \item A sequence of words and expressions and/or equations or;
    \item A sequence of only equations.
\end{itemize}
\end{definition}

\begin{definition}A \textbf{mathematical text} $\mathcal{\tau}$ is a sequence $\{\mu_1, \mu_2, \mu_3, ..., \mu_n\}$ of mathematical statements.

\end{definition}

In the mathematical text, words, expressions and equations, can be directly related through a relationship of \textit{definiendum} and \textit{definiens}, where an expression, the definiendum, is defined by a mathematical statement or part of a mathematical statement, the definiens is also used to determine the \textit{set of values} and properties associated with an expression.

\begin{definition}A \textbf{mathematical definiens} $\sigma_{\tau_i}$ is the set of tuples composed by  $e$  and the set of (part of) mathematical statements that declares and/or quantifies $e$ in the mathematical text $\tau_i$. Figure~\ref{fig:definiendum} presents an example where the \textit{definiens} and the \textit{definiendum} are highlighted. A definiendum can have more than one \textit{definiens}, for example, the expression ``$f: \mathbb{R}\to \mathbb{R}$'' is declared by the equation ``$f \left({x}\right) = a^x$'', and has the property ``\textit{real function}''. Therefore:

    \begin{equation*}
    \begin{aligned}
    \sigma_{\tau_{example}} = &  \{(``f: \mathbb{R}\to \mathbb{R}", ``f \left({x}\right) = a^x"), \\
          & (``f: \mathbb{R}\to \mathbb{R}", ``\text{real function}"), ...\}
    \end{aligned}
    \end{equation*}
    
    \begin{figure}
    \begin{tcolorbox}
    Let \mybox{$a \in \mathbb{R}_{>0}$} be a 
    \underline{strictly positive real number}.
    
    Let \mybox{$f: \mathbb{R}\to \mathbb{R}$} be the  \underline{real function} defined as:
    \underline{$f \left({x}\right) = a^x$}
    where \underline{$a^x$} denotes \underline{$a$ to the power of $x$}.
    
    Then \mybox{$f$} is \underline{convex}.
    \end{tcolorbox}
    \caption{Theorem with three definiendums and six definiens, where the content inside the boxes are definiendums and the underlined content are definiens.}
    \label{fig:definiendum}
    \end{figure}
    
\end{definition}

Different mathematical texts can also be related, since mathematical knowledge is often incremental, where one element depends on others. For example, in Figure~\ref{fig:definiendum}, in order to understand the meaning of the presented text, we need to understand the definition of a \textit{real function}, which is defined in another mathematical text.

\begin{definition}A \textbf{mathematical supporting definition} $\delta_{\tau_i}$ 
is the set of mathematical texts $\{\tau_j,\tau_k, \tau_l, ... \}$, where all elements in $\delta_{\tau_i}$ contains a definition of a concept presented in $\tau_i$. For example, the theorem in Figure~\ref{fig:definiendum} is connected to the mathematical text that defines what is a real function.
\end{definition}

\begin{definition}A \textbf{definition} $\mathcal{D}$ is composed by a 4-tuple $(\tau, c, \sigma_{\tau}, \delta_{\tau} )$, where $\tau$ is the definition text, $c$ is the set of categories that the definition belongs to, $\sigma_{\tau}$ is the set of definiens in the text and $\delta_{\tau}$ is the set definitions that is referenced in $\mathcal{D}$. If $\delta_{\tau}$ is empty, we call it an \textbf{atomic definition}.
\end{definition}

A mathematical proof is a particular mathematical text that tries to convince the reader that a specific hypothesis can lead to a conclusion~\cite{solow2002read}. Proofs often contain mathematical bindings. They can also be connected to other propositions, such as lemmas, theorems and corollaries, as we will define next.

\begin{definition}A \textbf{mathematical supporting proposition} $\omega_{\tau_i}$ is the set of propositions that helps support the argument proposed in the mathematical text $\tau_i$ of a proof. It is often used as an explanation for certain statements used for the construction of the proof. Figure~\ref{fig:proof} presents part of the proof, where the name of the supporting facts is highlighted. For example, the mathematical statement of Cauchy's Mean Theorem is a supporting fact for the proof shown. 

\begin{figure}
\begin{tcolorbox}[boxsep=-1.5mm]
Let $x, y \in \mathbb{R}$.

Note that, from \colorbox{cyan}{Power of Positive Real}
\colorbox{cyan}{Number is Positive: Real Number}:

$\forall t \in \mathbb{R}: a^t > 0$.

So:
\begin{flalign*}
a^{\left({x + y}\right) / 2} && = && \sqrt {a^{x + y} } && \text{(\footnotesize{\colorbox{cyan}{Exponent Combination Laws})}}\\
&& = && \sqrt {a^x a^y} && \text{(\footnotesize{\colorbox{cyan}{Exponent Combination Laws})}}\\
&& \le && \frac {a^x + a^y} 2 && \text{(\footnotesize{\colorbox{cyan}{Cauchy's Mean Theorem})}}\\
\end{flalign*}

\end{tcolorbox}
\caption{Example of part of a proof, where four mathematical supporting facts are present.}
\label{fig:proof}
\end{figure}
\end{definition}

\begin{definition}
The set of \textbf{premises} $\phi_{\tau_i}$ of a mathematical text $\tau_i$ of a proof is the set of supporting facts $\omega_{\tau_i}$ and the set of supporting definitions $\delta_{\tau_i}$, i.e., $\phi_{\tau_i} = \omega_{\tau_i} \cup \delta_{\tau_i}$.
\end{definition}

\begin{definition}A \textbf{mathematical proof} $\mathcal{P}$ is defined is composed by a tuple $(\tau, \phi_{\tau})$, where $\tau$ is the proof text and $\phi_{\tau}$ is the set of premises for $\mathcal{P}$.
\end{definition}

\theoremstyle{definition}
\begin{definition} A \textbf{theorem} $\mathcal{T}$ is composed by a tuple $(\tau, c, \sigma_{\tau}, \mathcal{P}$), where $\tau$ is the theorem's text, $c$ is the set of categories that the theorem belongs to, $\sigma_\tau$ is the set of definiens in the text, $P$ is the set of proofs for the theorem (one theorem can have more than one possible proof).
\end{definition}

\begin{definition}Similarly, we can define a \textbf{lemma} $\mathcal{L}$. $\mathcal{L}$ is composed by a  5-tuple $(\tau, c, \sigma_\tau, \mathcal{P}, t)$. With the addition of $t$, theorem where the lemma occurs.
\end{definition}

\theoremstyle{definition}
\begin{definition}
A \textbf{corollary} $\mathcal{C}$ is composed by a 5-tuple $(\tau, c, \sigma_{\tau}, \mathcal{P}, t)$, where $t$ is the theorem that derives $\mathcal{C}$.
\end{definition}

\section{Dataset Construction: NL-PS}

\begin{figure*}[!htbp]
    \centering
    \includegraphics[width=0.7\textwidth]{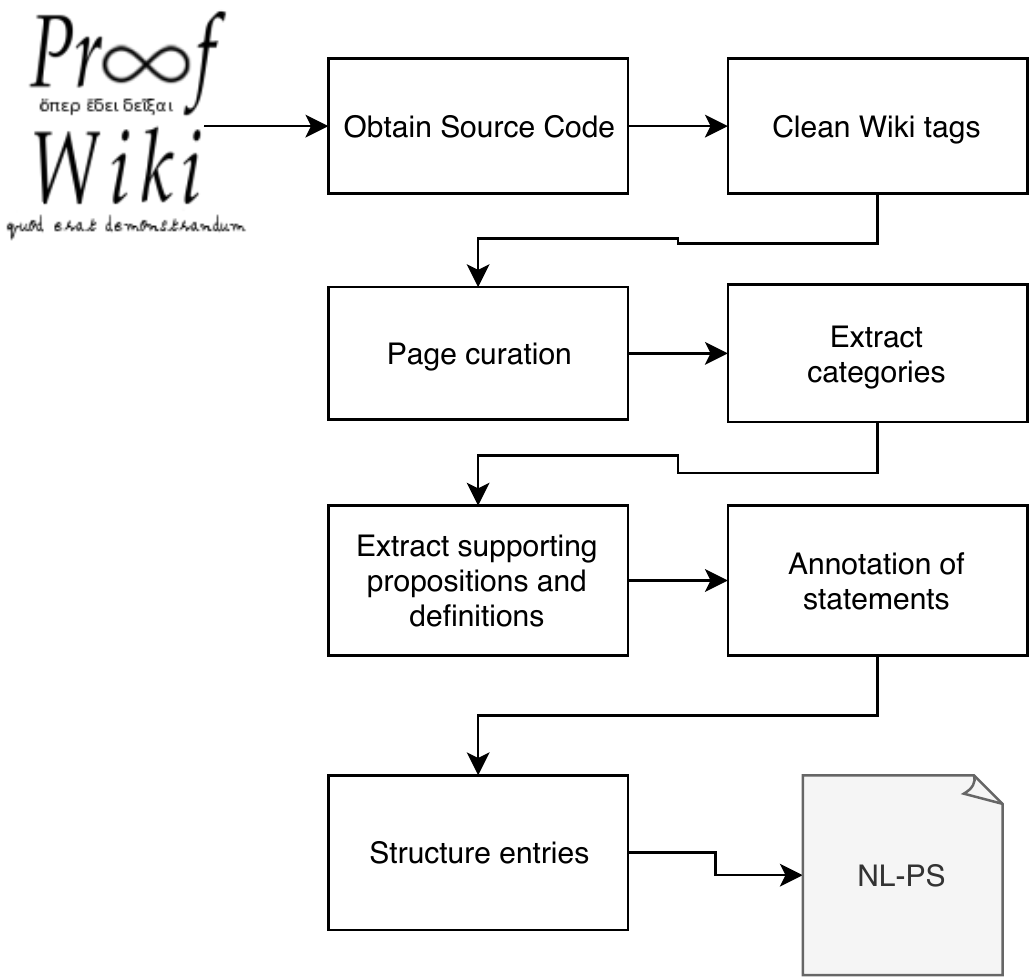}
    \caption{Pipeline used to build the NL-PS dataset.}
    \label{fig:nlps_build}
\end{figure*}

In this section, we present our dataset, NL-PS, and detail the steps we took in order to construct it. Our dataset is available as a set of JSON files in http://github.com/debymf/nl-ps. A summary of the process is presented in Figure~\ref{fig:nlps_build}.

\subsubsection*{Parsing the corpus}

The proposed dataset was extracted from the source code of ProofWiki. ProofWiki is an online compendium of mathematical proofs, with a goal to collect and classify mathematical proofs. ProofWiki contains links between theorems, definitions and axioms in the context of a mathematical proof, determining which dependencies are present. ProofWiki is manually curated by different collaborators; therefore, there are different styles of mathematical text and many elements cannot be extracted automatically.

\subsubsection*{Cleaning wiki tags}
\begin{figure}[!htbp]
    \centering
    \includegraphics[width=0.5\textwidth]{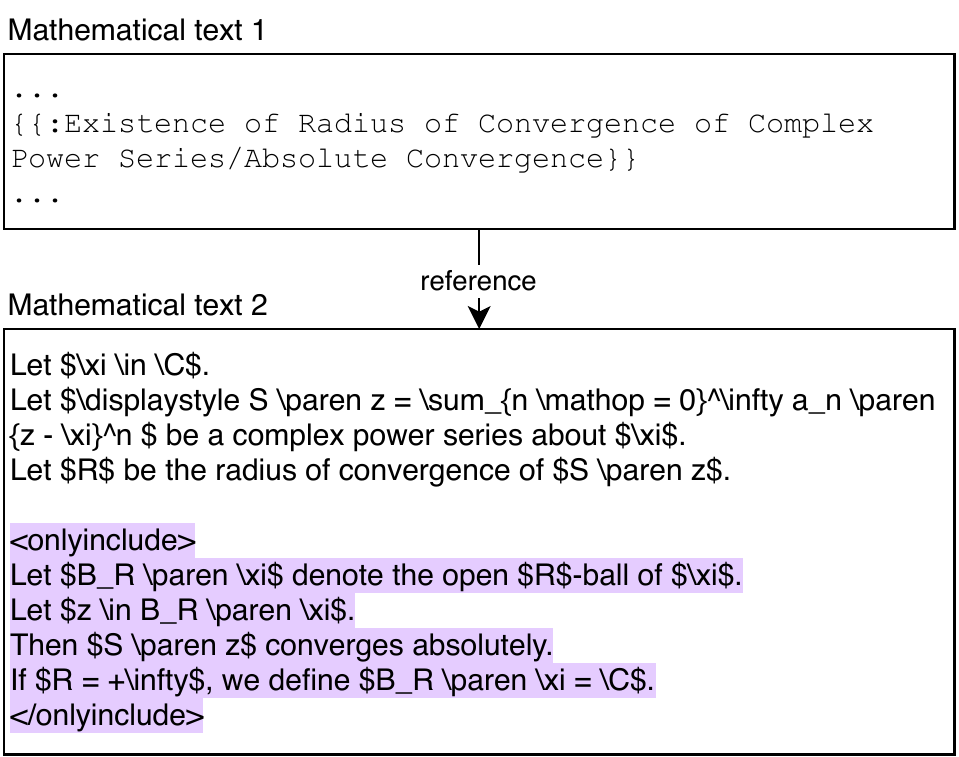}
    \caption{An example where Mathematical text 1 references a passage in Mathematical text 2 using the name of the passage to be referenced between curly brackets. Only the part highlighted is being referenced.}
    \label{fig:onlyinclude}
\end{figure}

ProofWiki has wikimedia tags; however, ProofWiki has also specific tags related to the mathematical domain. Therefore, we cannot use default wiki extraction tools. A bespoke tool was developed to comply with ProofWiki's tagging scheme. For example, there is a particular tag for referring to another mathematical text, using passages from other texts in order to support a claim (Figure~\ref{fig:onlyinclude}).

\subsubsection*{Proof curation}
Several pages in ProofWiki are not directly related to mathematical propositions or definitions, such as users pages, help pages, and pages about specific talks. We manually analysed the pages and removed the ones that are not definitions, lemmas, theorems or corollaries. Some pages also contained tags to indicate that 

\subsubsection*{Extraction of categories}
ProofWiki has associated categories for each page. However, the categories are not harmonised across definitions and propositions. We merged different categories that belonged to the same mathematical branch and selected the categories that contained at least 100 different entries.
The categories selected are: Analysis, Set Theory, Number Theory, Abstract Algebra, Topology, Algebra, Relation Theory, Mapping Theory, Real Analysis, Geometry, Metric Spaces, Linear Algebra, Complex Analysis, Applied Mathematics, Order Theory, Numbers, Physics, Group Theory, Ring Theory, Euclidean Geometry, Class Theory, Discrete Mathematics, Plane Geometry and Units of Measurement

\subsubsection*{Extracting supporting facts}

\begin{figure}[!htbp]
    \centering
    \includegraphics[width=0.5\textwidth]{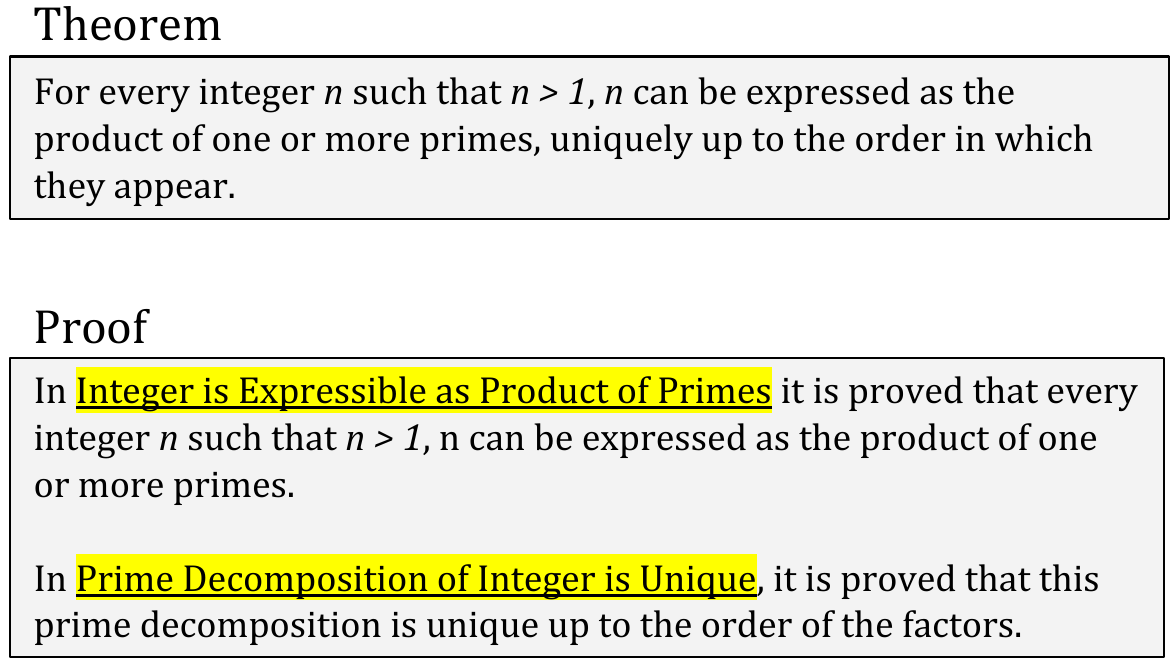}
    \caption{Example of supporting propositions for a theorem.}
    \label{fig:theoremandproof}
\end{figure}

The pages in ProofWiki are connected using hyperlinks. We leverage this structure to extract supporting propositions and supporting definitions. From the definition mathematical text, we extract the hyperlinks connecting to other definitions and these links are the supporting definitions. From the mathematical text of proofs, we extract the hyperlinks to other propositions and we consider these as supporting propositions. For example, Figure~\ref{fig:theoremandproof} presents a theorem and its respective proof. The proof contain links (highlighted) to other propositions, these are supporting propositions needed in order to support the proof.

\subsubsection*{Annotating mathematical text}

The entries in ProofWiki are often divided in sections, for NL-PS, we are only interested in the sections that present a definition, a proposition or a proof. Proofs were curated (combining manual and automatic annotation) to contain only mathematic discourse, removing satellite discourse such as \textit{Historical Notes}. Because some propositions can be proved in different ways, we also annotated the different proofs which can be found inside one single page.

\subsubsection*{Structuring the entries}
Finally, the dataset is structured as follows:
\begin{itemize}
    \item Definitions entries are composed by a mathematical text and a set of supporting definitions.
    \item Lemmas and Theorems have a mathematical text, a proof and a set of premises.
    \item Corollaries are composed by a mathematical text, a proof, a set of premises and the theorem that derives the corollary.
\end{itemize}

\section{Dataset Analysis}

NL-PS has a total of 20,401 different entries, composed of definitions, lemmas, corollaries and theorems, as shown in Table~\ref{tab:categories}.

\begin{table}[!htbp]
\centering

\label{tab:categories}
\begin{tabular}{@{}lr@{}}
\toprule
\multicolumn{1}{c}{Type} & \multicolumn{1}{c}{Number of entries} \\ \midrule
Definitions              & 5,633                                 \\
Lemmas                   & 327                                   \\
Corollaries              & 292                                   \\
Theorems                 & 14,149                                \\
Total                    & 20,401                                \\ \bottomrule
\end{tabular}
\caption{Types of mathematical documents in NL-PS}
\end{table}

\begin{figure}[!htbp]
    \centering
    \includegraphics[width=0.5\textwidth]{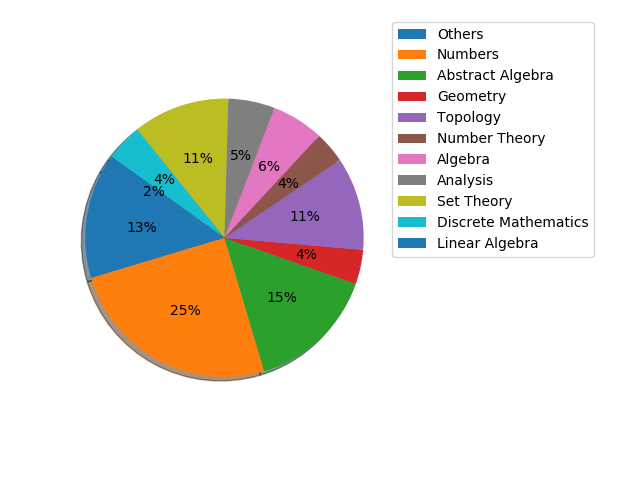}
    \caption{Distribution of documents per category in the dataset.}
    \label{fig:categories_entries}
\end{figure}

Figure~\ref{fig:categories_entries} presents the distribution of different categories in the dataset.

\begin{figure}[!htbp]
  \includegraphics[width=\linewidth]{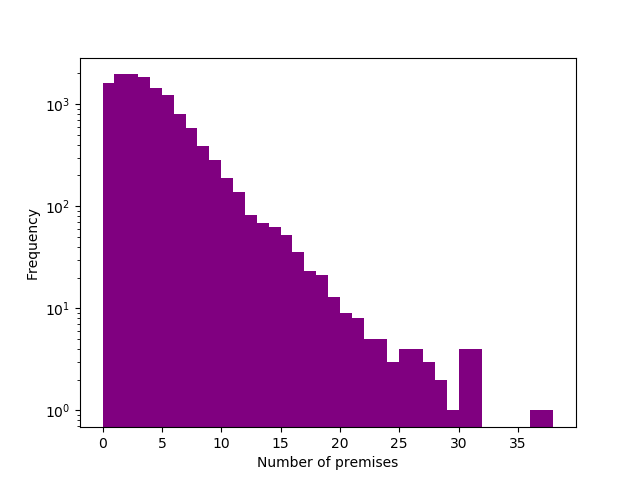}
  \caption{Distribution of the number of premises in the ProofWiki corpus.}
  \label{fig:chart_premises}
\end{figure}

Figure~\ref{fig:chart_premises} presents a histogram with the frequency of the different number of premises. We can observe that the statements usually have a small number of premises, with $8,046$ statements containing between one and five premises. The highest number of premises for one theorem is $32$ (text for the theorem ``The Sorgenfrey line is Lindel{\"o}f.'').

\begin{figure}[!htbp]
  \includegraphics[width=\linewidth]{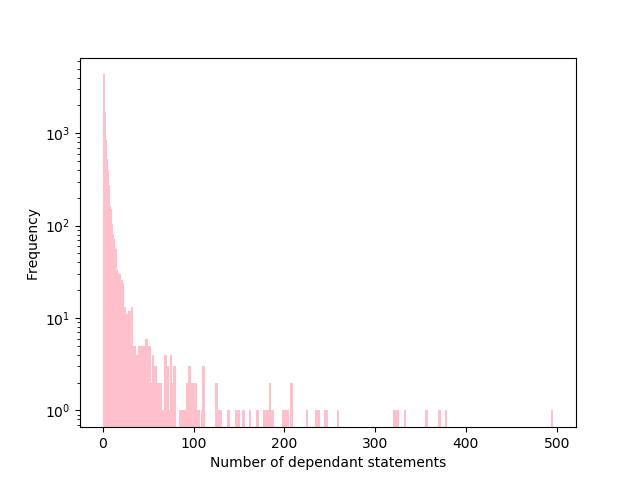}
  \caption{Number of times a statement is referred as a premise.}
  \label{fig:chart_premises_of}
\end{figure}

Similarly, the histogram in Figure~\ref{fig:chart_premises_of} shows the frequency of the different number of dependencies. 

We also computed how many times each statement is used as a premise, and we observed that most of the statements are used as dependencies for only a small subset of premises. A total of $6,866$ statements has between one and three dependants. On average, statements contain a total of $289$ symbols (characters and mathematical symbols). The specific number of tokens will depend on the type of tokenisation used for the mathematical symbols. 

We can also represent the connections (premises) between different mathematical texts as a graph. This graph has a total of 14,393 nodes (the number of nodes is smaller than the number entries, since some of the entries are disconnected, and we do not consider those for the graph) and 34,874 edges.  

\begin{figure}[!htbp]
    \centering
    \includegraphics[width=0.35\textwidth]{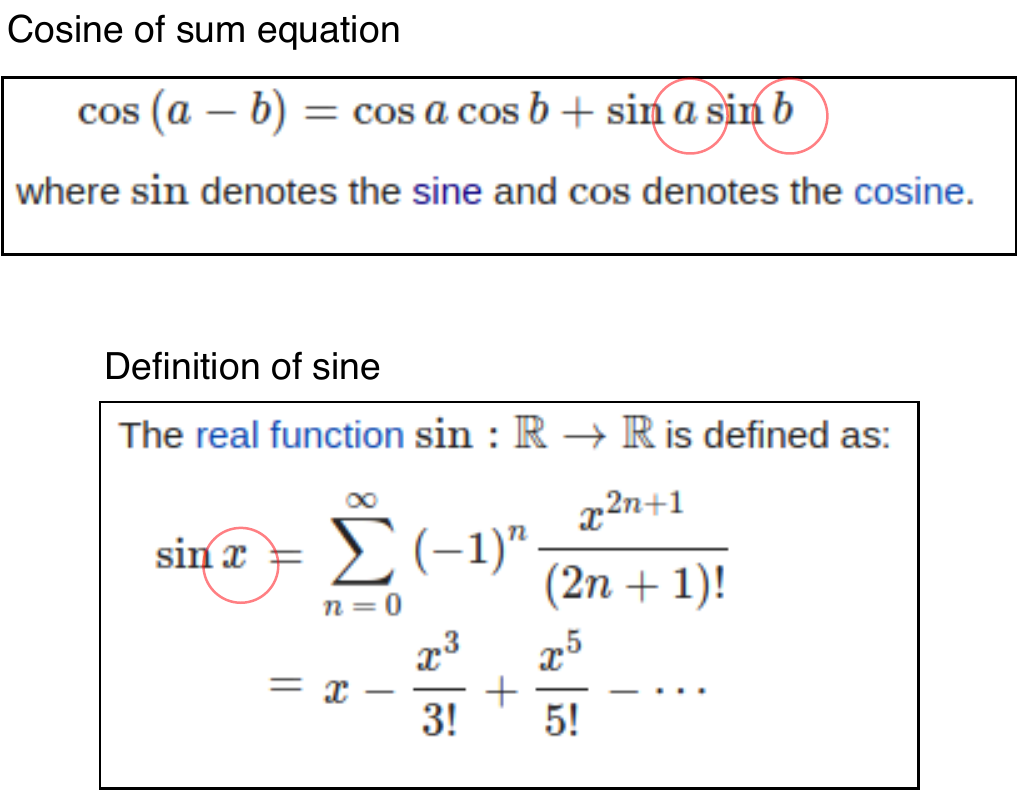}
    \caption{Variables with different symbols, but referring to the same set.}
    \label{fig:variables}
\end{figure}

The dataset provides a specific semantic modelling challenge for natural language processing as it requires specific tokenization, co-reference resolution and the modelling of specific discourse structures tailored towards  mathematical text. One crucial challenge is how to resolve the semantics of variables in mathematical expressions, which requires a particular binding method. As shown in Figure~\ref{fig:variables}, variables that refer to the same set can often have different names. For example, in the definition of sine, the variable being used is $x$, but $a$ and $b$ refers to the same set. Basically, variables serve as a mathematical alternative to anaphora~\cite{ganesalingam2013language}. 

\section{Experiments}

In order to identify the challenges of the task of natural premise selection using NL-PS, we performed initial experiments using two baselines: TF-IDF and PV-DBOW~\cite{le2014distributed}. We use both techniques to create vector representations for all the mathematical texts. Then compute the cosine similarity between each entry and rank the results by proximity. We then compute the Mean Average Precision (MAP) for each baseline, ranking all possible premises, computed as:
\begin{equation*}
    \operatorname{MAP} = \frac{\sum_{i=1}^N \operatorname{AveP(\mathcal{\tau}_i)}}{N}
\end{equation*}

where $N$ is the total number of documents, $\mathcal{\tau}_i$ is the $i$-th mathematical text and AveP is the average precision.

Table~\ref{tab:my-table1} presents the initial results. We compare three different types of tokenisations for the mathematical elements. Initially, we treat the expressions and equations as single tokens; for example, the expression ``$x+y+z$'' would be considered a single word. We also considered tokenised expressions, tokenising operations and operators, the expression ``$x+y+z$'' would be tokenised as [`$x$',`$+$',`$y$',`$+$',`$z$']. Finally, we tokenise the whole text as a sequence of characters. We run PV-DBOW with the default parameters, comparing different sizes of embeddings, with the best results obtained with an embedding size of 100.  

From these initial results, we can conclude that the task is semantically non-trivial and cannot be solved with retrieval strategies such as lexical overlap. We can also notice that we obtain better results when we tokenise the expressions, hinting that the elements inside the expressions have semantic properties that are relevant for determining the relevant premises. For the following experiments, we are using the tokenised expressions and PV-DBOW with an embedding size of 100.

\begin{table}[!htbp]
\centering

\label{tab:my-table1}

\begin{tabular}{@{}lllll@{}}
\toprule
\multicolumn{1}{c}{}  & \multicolumn{1}{c}{TFIDF} & \multicolumn{3}{c}{PV-DBOW}                                                \\
\multicolumn{1}{c}{}  & \multicolumn{1}{c}{}      & \multicolumn{1}{c}{50} & \multicolumn{1}{c}{100} & \multicolumn{1}{c}{200} \\ \midrule
Expression as words   & 0.073                     & 0.048                  & 0.051                   & 0.046                   \\
Tokenised expressions & \textbf{0.089}                     & \textbf{0.069}                  & \textbf{0.073}                   & \textbf{0.072}                   \\
Char level            & 0.051                     & 0.059                  & 0.065                   & 0.061                   \\ \bottomrule
\end{tabular}%
\caption{MAP results for TF-IDF and PV-DBOW comparing tokenisation of expressions. We compare the results for PV-DBOW for different dimension values.}
\end{table}

In Table~\ref{tab:my-table2} we compare the results for different sizes of the dataset. We consider the full dataset and three different subsets with different categories. We can notice that for smaller datasets, both baselines perform better. This result was expected since with smaller datasets there are less possible premises, and elements from the same categories tend to be more uniform between themselves.

\begin{table}[!htbp]
\centering

\label{tab:my-table2}

\begin{tabular}{@{}lrr@{}}
\toprule
               & TFIDF & PV-DBOW \\ \midrule
All Categories &   0.089     &         0.076                          \\
Algebra (1,241)       &     0.183   &            0.177                    \\
Analysis    (1,102)   &    0.191    &            0.212                      \\
Number Theory (741) &   0.242     &        0.188 \\ \bottomrule
\end{tabular}%
\caption{Comparing results for different categories (the number between parenthesis indicates the number of entries for that category).}
\end{table}

We can also consider the fact that the premises are transitive, i.e., if one a mathematical text $\tau_i$ has a premise $x$ and a mathematical text $\tau_j$ has $\tau_i$ as a premise, then $x$ should also be a premise of $\tau_j$. In this case, the task becomes even more challenging, as we present in Table~\ref{tab:my-table3}, where we consider the transitivity with two and three hops of distance. From the results, we notice that the more hops needed to obtain the premise, the worse our baselines perform.

\begin{table}[!htbp]
\centering

\label{tab:my-table3}

\begin{tabular}{@{}lrr@{}}
\toprule
               & TFIDF & PV-DBOW \\ \midrule
1-hop premises &   0.089     &      0.073                          \\
2-hop premises &    0.052    &         0.047                         \\
3-hop premises &    0.038    &          0.031                        \\ \bottomrule
\end{tabular}%
\caption{Comparing number of hops needed for obtaining premises.}
\end{table}

We also verify on how state-of-the-art embedding models perform with such specific dataset. BERT~\cite{devlin2018bert} is reported to have performed in different NLP tasks, including understanding numeracy~\cite{wallace2019nlp}. 

In order to use BERT, we formulate the problem as a pairwise relevance classification problem, where we aim to classify if one mathematical text is connected to another. We do not perform any pre-processing for the expressions.

For this experiment, we used the pre-trained BERT model \textit{bert-base-uncased} and SciBERT~\cite{beltagy2019scibert} model \textit{scibert-scivocab-uncased}, fine-tuning for our task with a sequence classifier, adding a linear layer on top of the transformer vectors. The results are presented in Table~\ref{tab:bert}. Even though BERT is not pre-trained using a mathematical corpus, it performs better than TF-IDF and PV-DBOW. SciBERT perform slightly better than BERT, since it was trained in a scientific corpus, but not in a mathematical corpus. This hints that BERT trained from scratch in a mathematical corpus could have even better results, however, this is outside the scope of this work.

\begin{table}[]
\centering

\label{tab:bert}

\begin{tabular}{@{}ll@{}}
\toprule
Model   & MAP   \\ \midrule
SciBERT & 0.383 \\
BERT    & 0.377 \\ \bottomrule
\end{tabular}%
\caption{Results for BERT and SciBERT.}
\end{table}

\section{Conclusion}

In this paper we proposed a new task for mathematical language processing: natural language premise selection. We also made a new dataset available for the evaluation of the task and we analysed how the dataset works with the task on different baselines.

From our experiments we identified that handling mathematical symbols are crucial for solving the task, taking into consideration more specific semantics of operators and variables: such semantics are not captured using PV-DM  or  BERT. This  provides  evidence on the need for specific embeddings and representation for mathematical  formulas and discourse,  which  could  most  certainly  improve  the  prediction  of  future  work  in  the  natural language premise selection task.

We also identify that the task becomes more challenging when we consider that the premises are transitive, suggesting that the task could benefit from  graph-based representations.

Our dataset can be used in a different set of natural mathematical reasoning tasks, aiding researchers on the creation of mechanisms for improving the way machines understand mathematical text.

\section{Acknowledgements}

The authors would like to thank the anonymous reviewers for the constructive feedback.

\section{Bibliographical References}
\label{main:ref}

\bibliographystyle{lrec}
\bibliography{lrec2020W-xample}

\begin{thebibliography}{}

\bibitem[\protect\citename{Aizawa \bgroup et al.\egroup }2014]{aizawa2014ntcir}
Aizawa, A., Kohlhase, M., Ounis, I., and Schubotz, M.
\newblock (2014).
\newblock Ntcir-11 math-2 task overview.
\newblock In {\em NTCIR}, volume~11, pages 88--98. Citeseer.

\bibitem[\protect\citename{Alama \bgroup et al.\egroup
  }2014a]{alama2014premise}
Alama, J., Heskes, T., K{\"u}hlwein, D., Tsivtsivadze, E., and Urban, J.
\newblock (2014a).
\newblock Premise selection for mathematics by corpus analysis and kernel
  methods.
\newblock {\em Journal of Automated Reasoning}, 52(2):191--213.

\bibitem[\protect\citename{Alama \bgroup et al.\egroup }2014b]{Alama2014}
Alama, J., Heskes, T., K{\"u}hlwein, D., Tsivtsivadze, E., and Urban, J.
\newblock (2014b).
\newblock Premise selection for mathematics by corpus analysis and kernel
  methods.
\newblock {\em Journal of Automated Reasoning}, 52(2):191--213, Feb.

\bibitem[\protect\citename{Beltagy \bgroup et al.\egroup
  }2019]{beltagy2019scibert}
Beltagy, I., Lo, K., and Cohan, A.
\newblock (2019).
\newblock Scibert: A pretrained language model for scientific text.
\newblock In {\em Proceedings of the 2019 Conference on Empirical Methods in
  Natural Language Processing and the 9th International Joint Conference on
  Natural Language Processing (EMNLP-IJCNLP)}, pages 3606--3611.

\bibitem[\protect\citename{Blanchette \bgroup et al.\egroup
  }2016]{blanchette2016hammering}
Blanchette, J.~C., Kaliszyk, C., Paulson, L.~C., and Urban, J.
\newblock (2016).
\newblock Hammering towards qed.
\newblock {\em Journal of Formalized Reasoning}, 9(1):101--148.

\bibitem[\protect\citename{Chaganty and Liang}2016]{chaganty2016much}
Chaganty, A. and Liang, P.
\newblock (2016).
\newblock How much is 131 million dollars? putting numbers in perspective with
  compositional descriptions.
\newblock In {\em Proceedings of the 54th Annual Meeting of the Association for
  Computational Linguistics (Volume 1: Long Papers)}, pages 578--587.

\bibitem[\protect\citename{Cramer \bgroup et al.\egroup
  }2009]{cramer2009naproche}
Cramer, M., Fisseni, B., Koepke, P., K{\"u}hlwein, D., Schr{\"o}der, B., and
  Veldman, J.
\newblock (2009).
\newblock The naproche project controlled natural language proof checking of
  mathematical texts.
\newblock In {\em International Workshop on Controlled Natural Language}, pages
  170--186. Springer.

\bibitem[\protect\citename{Devlin \bgroup et al.\egroup }2019]{devlin2018bert}
Devlin, J., Chang, M.-W., Lee, K., and Toutanova, K.
\newblock (2019).
\newblock Bert: Pre-training of deep bidirectional transformers for language
  understanding.
\newblock In {\em Proceedings of the 2019 Conference of the North American
  Chapter of the Association for Computational Linguistics: Human Language
  Technologies, Volume 1 (Long and Short Papers)}, pages 4171--4186.

\bibitem[\protect\citename{F{\"a}rber and
  Kaliszyk}2015]{10.1007/978-3-319-24246-0_20}
F{\"a}rber, M. and Kaliszyk, C.
\newblock (2015).
\newblock Random forests for premise selection.
\newblock In Carsten Lutz et~al., editors, {\em Frontiers of Combining
  Systems}, pages 325--340, Cham. Springer International Publishing.

\bibitem[\protect\citename{Ganesalingam and Gowers}2017]{ganesalingam2017fully}
Ganesalingam, M. and Gowers, W.~T.
\newblock (2017).
\newblock A fully automatic theorem prover with human-style output.
\newblock {\em Journal of Automated Reasoning}, 58(2):253--291.

\bibitem[\protect\citename{Ganesalingam}2013]{ganesalingam2013language}
Ganesalingam, M.
\newblock (2013).
\newblock The language of mathematics.
\newblock In {\em The Language of Mathematics}, pages 17--38. Springer.

\bibitem[\protect\citename{Gao \bgroup et al.\egroup
  }2017]{DBLP:journals/corr/GaoJYYYT17}
Gao, L., Jiang, Z., Yin, Y., Yuan, K., Yan, Z., and Tang, Z.
\newblock (2017).
\newblock {Preliminary Exploration of Formula Embedding for Mathematical
  Information Retrieval: can mathematical formulae be embedded like a natural
  language?}
\newblock {\em CIKM 2017 Workshop on Interpretable Data Mining (IDM)}.

\bibitem[\protect\citename{Gauthier and
  Kaliszyk}2015]{Gauthier:2015:PSE:2676724.2693173}
Gauthier, T. and Kaliszyk, C.
\newblock (2015).
\newblock Premise selection and external provers for hol4.
\newblock In {\em Proceedings of the 2015 Conference on Certified Programs and
  Proofs}, CPP '15, pages 49--57, New York, NY, USA. ACM.

\bibitem[\protect\citename{Greiner-Petter \bgroup et al.\egroup
  }2019]{Greiner-Petter2019WhyYet}
Greiner-Petter, A., Ruas, T., Schubotz, M., Aizawa, A., Grosky, W., and Gipp,
  B.
\newblock (2019).
\newblock {Why Machines Cannot Learn Mathematics, Yet}.
\newblock {\em 4th BIRNDL workshop at 42nd SIGIR}.

\bibitem[\protect\citename{Huang \bgroup et al.\egroup }2016]{huang2016well}
Huang, D., Shi, S., Lin, C.-Y., Yin, J., and Ma, W.-Y.
\newblock (2016).
\newblock How well do computers solve math word problems? large-scale dataset
  construction and evaluation.
\newblock In {\em Proceedings of the 54th Annual Meeting of the Association for
  Computational Linguistics (Volume 1: Long Papers)}, volume~1, pages 887--896.

\bibitem[\protect\citename{Irving \bgroup et al.\egroup
  }2016]{irving2016deepmath}
Irving, G., Szegedy, C., Alemi, A.~A., Een, N., Chollet, F., and Urban, J.
\newblock (2016).
\newblock Deepmath-deep sequence models for premise selection.
\newblock In {\em Advances in Neural Information Processing Systems}, pages
  2235--2243.

\bibitem[\protect\citename{Krstovski and
  Blei}2018]{Krstovski2018EquationEmbeddings}
Krstovski, K. and Blei, D.~M.
\newblock (2018).
\newblock {Equation Embeddings}.

\bibitem[\protect\citename{Le and Mikolov}2014]{le2014distributed}
Le, Q. and Mikolov, T.
\newblock (2014).
\newblock Distributed representations of sentences and documents.
\newblock In {\em International conference on machine learning}, pages
  1188--1196.

\bibitem[\protect\citename{Mikolov \bgroup et al.\egroup
  }2013]{mikolov2013distributed}
Mikolov, T., Sutskever, I., Chen, K., Corrado, G.~S., and Dean, J.
\newblock (2013).
\newblock Distributed representations of words and phrases and their
  compositionality.
\newblock In {\em Advances in neural information processing systems}, pages
  3111--3119.

\bibitem[\protect\citename{Solow}2002]{solow2002read}
Solow, D.
\newblock (2002).
\newblock How to read and do proofs an introduction to mathematical thought
  processes.

\bibitem[\protect\citename{Wallace \bgroup et al.\egroup }2019]{wallace2019nlp}
Wallace, E., Wang, Y., Li, S., Singh, S., and Gardner, M.
\newblock (2019).
\newblock Do nlp models know numbers? probing numeracy in embeddings.
\newblock In {\em Proceedings of the 2019 Conference on Empirical Methods in
  Natural Language Processing and the 9th International Joint Conference on
  Natural Language Processing (EMNLP-IJCNLP)}, pages 5310--5318.

\bibitem[\protect\citename{Wang \bgroup et al.\egroup }2018]{wang2018first}
Wang, Q., Kaliszyk, C., and Urban, J.
\newblock (2018).
\newblock First experiments with neural translation of informal to formal
  mathematics.
\newblock In {\em International Conference on Intelligent Computer
  Mathematics}, pages 255--270. Springer.

\bibitem[\protect\citename{Warner}1990]{warner1990modern}
Warner, S.
\newblock (1990).
\newblock {\em Modern algebra}.
\newblock Courier Corporation.

\bibitem[\protect\citename{Zinn}2003]{zinn2003computational}
Zinn, C.
\newblock (2003).
\newblock A computational framework for understanding mathematical discourse.
\newblock {\em Logic Journal of IGPL}, 11(4):457--484.

\end{thebibliography}


\end{document}